\documentclass[runningheads]{llncs}
\usepackage{graphicx}
\usepackage{todonotes}
\usepackage{booktabs}
\usepackage{multirow}
\usepackage{hyperref}
\usepackage{subcaption}
\usepackage[export]{adjustbox}
\usepackage{color,ulem}
\usepackage{array}
\usepackage{xcolor}
\definecolor{blue}{HTML}{4472C4}
\definecolor{green}{HTML}{70AD47}
\definecolor{orange}{HTML}{ED7D31}
\definecolor{purple}{HTML}{7030A0}
\definecolor{yellow}{HTML}{FFC000}
\definecolor{teal}{HTML}{00CC99}
\definecolor{pink}{HTML}{CC0099}

\newcommand{\cbox}[1]{\colorbox{#1}{\phantom{0}}}
\newcommand\blueuline{\bgroup\markoverwith{\textcolor{blue}{\rule[-0.5ex]{2pt}{0.4pt}}}\ULon}
\newcommand\greenuline{\bgroup\markoverwith{\textcolor{green}{\rule[-0.5ex]{2pt}{0.4pt}}}\ULon}
\newcommand\orangeuline{\bgroup\markoverwith{\textcolor{orange}{\rule[-0.5ex]{2pt}{0.4pt}}}\ULon}
\newcommand\purpleuline{\bgroup\markoverwith{\textcolor{purple}{\rule[-0.5ex]{2pt}{0.4pt}}}\ULon}
\newcommand\yellowuline{\bgroup\markoverwith{\textcolor{yellow}{\rule[-0.5ex]{2pt}{0.4pt}}}\ULon}
\newcommand\tealuline{\bgroup\markoverwith{\textcolor{teal}{\rule[-0.5ex]{2pt}{0.4pt}}}\ULon}
\newcommand\pinkuline{\bgroup\markoverwith{\textcolor{pink}{\rule[-0.5ex]{2pt}{0.4pt}}}\ULon}

\begin{document}
\title{SIMARA: a database for key-value information extraction from full-page handwritten documents}
\titlerunning{SIMARA: a database for key-value information extraction from full pages}
%
\author{Solène Tarride\inst{1}\orcidID{0000-0001-6174-9865} \and
Mélodie Boillet\inst{1,2}\orcidID{0000-0002-0618-7852} \and
Jean-François Moufflet\inst{3} \and
Christopher Kermorvant\inst{1,2}\orcidID{0000-0002-7508-4080} 
}
\authorrunning{S. Tarride et al.}
%
\institute{TEKLIA, Paris, France \and
LITIS, Normandy University, Rouen, France \and
Archives nationales, Paris, France}
\maketitle              
%


\begin{abstract}

We propose a new database for information extraction from historical handwritten documents. The corpus includes 5,393 finding aids from six different series, dating from the 18th-20th centuries. Finding aids are handwritten documents that contain metadata describing older archives. They are stored in the National Archives of France and are used by archivists to identify and find archival documents. 
Each document is annotated at page-level, and contains seven fields to retrieve. The localization of each field is not available in such a way that this dataset encourages research on segmentation-free systems for information extraction.
We propose a model based on the Transformer architecture trained for end-to-end information extraction and provide three sets for training, validation and testing, to ensure fair comparison with future works. The database is freely accessible\footnote{\url{https://zenodo.org/record/7868059}}.

\end{abstract}
\section{Introduction}
After transforming information retrieval on the web, machine learning and deep learning techniques are now commonly used in archives\cite{Colavizza2021}. The major difference with the natively digital data of the web or traditional information systems, is that the majority of documents kept in archives are physical documents. These documents have been massively digitised in recent years, but only a few percent of the total collections have been digitised and as the volumes are colossal, complete digitisation of the archives is not envisaged. It is therefore necessary to select the documents to be digitised, according to their usefulness. Among the documents kept by the archives, finding aids play a special role. These documents have been produced by the archive services for decades to facilitate access to the information contained in the documents. They are the main gateway to the mass of documents held and are therefore consulted as a priority for any research. Their digitisation and conversion to digital format of the indexing information they contain are therefore essential to improve the service provided to users.

While the automatic processing of original historical documents usually seeks to produce a textual transcription of their content\cite{muehlberger2019}, the same cannot be said of finding aids: for the latter, the automatic processing must make it possible to extract the indexing information they contain in a typed and standardised form, so that it can be inserted into a database. The processing is therefore more about extracting textual information than about pure transcription. The automatic processing chains for these documents must therefore include, in addition to the traditional stages of document analysis and handwriting recognition, information or entity (named entity) extraction \cite{Cunhal2022} and normalisation stages. 

Paradoxically, while finding aids are central to accessing archival information, they have hardly been studied in the document analysis community: no such database is publicly available and document recognition systems are not evaluated on this type of document. It is to fill this gap that we make available to the community the SIMARA database described in this article. In addition, we also wish to introduce to the community a new type of task for automatic processing models: a task for extracting key-value information from historical documents by processing the whole page. This task differs from traditional transcription or named entity extraction tasks in the following aspects :
\begin{itemize}
    \item the entire document must be considered to perform the information extraction (full page)
    \item among all the text contained in the document, only a part has to be extracted
    \item each extracted information must be typed, it is a key (type)-value association
    \item the position of the information on the document is not necessary
    \item the extracted information (value) must be normalized to satisfy the constraints of the database it will enrich.
\end{itemize}

In this paper, we propose two main contributions. First, we present SIMARA, a new dataset of finding aids images, annotated with the text transcriptions and their corresponding fields at page-level. We also present an end-to-end baseline model trained on SIMARA, which performs, in a single forward step, the extraction of the target key-value information.

This paper is organized as follows. In Section 2, we review the different datasets related to SIMARA as well as the recently proposed models for extracting text and entities from full-page images. Section 3 presents the SIMARA dataset and the different series in detail. Finally, in Section 4, we present first experiments and results of training automatic models for key-value information extraction from SIMARA images.



\section{Related work}

The SIMARA dataset addresses a rather different task compared to what can be found in the current literature. The challenge is to recognize handwritten text from full pages, but also the fields (key) corresponding to each piece of text (value). 
These fields are at the frontier between named entities and layout-based tokens. 

Many public datasets have been introduced for historical document image recognition. All these datasets have been listed and grouped by subtask through the tremendous work carried out by Nikolaidou et al. \cite{Nikolaidou2022} in their systematic review. 

In this section, we present some of the datasets that are the most similar to SIMARA. We also detail the various methods that have been proposed for end-to-end handwriting recognition from full pages, with or without entity extraction.

\subsection{Datasets}
Many datasets have been proposed for automatic handwriting recognition but, to our knowledge, none contain digitised finding aids. Generally, datasets can be classified into two main types according to their level of annotation:

\subsubsection{Datasets designed for text-line recognition:}

several datasets were designed for handwriting recognition from line images. As a result, the main objective is to recognize text from pre-segmented text lines.
For example, the Norhand database \cite{NorHand} contains documents from the 19th-20th centuries written in Norwegian, but only include text line images and their corresponding transcriptions.
Similarly, the IAM \cite{IAM} database have been designed for text recognition from line images. Most algorithms evaluated on this database are trained on text-lines, although full pages are available and reading order can be retrieved easily.

\subsubsection{Datasets designed for full page recognition:}

other datasets include full page images and annotations, including transcriptions and detailed layout information. This information is generally encoded in an XML PAGE ground truth file.
For example, the following datasets provide complete ground truth annotation at page-level: READ 2016 (German, 15-19th century) \cite{READ2016}, Digital Peter (Russian, 18th century) \cite{DigitalPeter}, ESPOSALLES (Catalan, 18th century) \cite{iehhr2017}, HOME-ALCAR (Latin, 12-14th centuries) \cite{HOME-ALCAR}, and POPP (French, 19-20th centuries) \cite{Constum-POPP}. 

Some of them also include named entity annotations. 
In ESPOSALLES \cite{iehhr2017}, two labels are associated to each word: a \textit{category} and a \textit{person}. As a result, this database allows training end-to-end models on full pages for handwriting recognition and named entity recognition. 
In POPP \cite{Constum-POPP}, there are no explicit named entities, but they can be retrieved using the column separator token included in transcriptions. 
Finally, in HOME-ALCAR \cite{HOME-ALCAR}, two named entities (place and person) are provided. 

Although full page-level annotations are available for these datasets, most researchers focus on text line detection or text line recognition.

\subsection{Models for end-to-end page recognition}

 In this section, we present automatic methods for end-to-end handwritten page recognition, with or without entity extraction. We focus only on systems that can handle full pages for documents.
 
Two main approaches have been proposed so far. First, \textit{segmentation-based} methods combine multiple models, each designed to tackle a specific task: text line detection, text recognition, and optionally named entity recognition. This type of approach is very popular as it is simpler to implement. 
The other approach consists in performing \textit{segmentation-free} handwriting recognition, and optionally named entity recognition, directly from full pages. 

\subsubsection{Segmentation-based models}

Constum et al. \cite{Constum-POPP} perform table detection from double page scans, page classification, text line detection and handwriting recognition on French census images from the POPP dataset. Similarly, Tarride et al. \cite{tarride2023} propose a complete workflow for end-to-end recognition of parish registers from Quebec. Their workflow includes page classification, text line detection, text recognition, and named entity recognition.

\subsubsection{Segmentation-free models}

Start Attend Read (SAR)  was the first HTR model proposed for paragraph recognition. The model is an attention-based CRNN-CTC network that attends over characters and manages to read multiple lines. 
Start Follow Read (SFR) \cite{Wigington2018-start-follow-read} combines a Region Proposal Network (RPN) combined with a CRNN-CTC network. As a result, it jointly learns line detection and recognition.
OrigamiNet \cite{OrigamiNet} is a CRNN-CTC network that unfolds multi-line features into a single line feature. 
Vertical Attention Network (VAN) \cite{VAN} is an attention-based model composed of an FCN encoder, a hybrid attention module, and a recurrent decoder. This model works at line or paragraph level and takes advantage of line breaks in the transcription to attend over line features.
All these models are able to recognize text from single pages or paragraphs, but cannot deal with multi-column text-blocks. 
The Document Attention Network (DAN) \cite{DAN} is the first model that can effectively process full documents, including documents featuring multiple columns and complex reading order. It is based on the Transformer architecture and jointly learns characters and special tokens that represent layout information. This model reaches state-of-the-art results on RIMES \cite{RIMES} and READ \cite{READ2016}
Finally, Rouhou et al. \cite{Rouhou2021-IEHHR-Transformer} introduce a Transformer model for combined HTR and NER on records on the ESPOSALLES database, where named entities are also represented by special tokens. They show that extracting information directly from records allows the model to benefit from more contextual information compared to a model trained on text lines. This model still requires record segmentation, but could potentially be applied to full pages.

\section{The SIMARA dataset} \label{section_simara_dataset}

\begin{figure}
     \centering
     \begin{subfigure}[b]{0.49\textwidth}
         \includegraphics[height=0.6\textwidth, center]{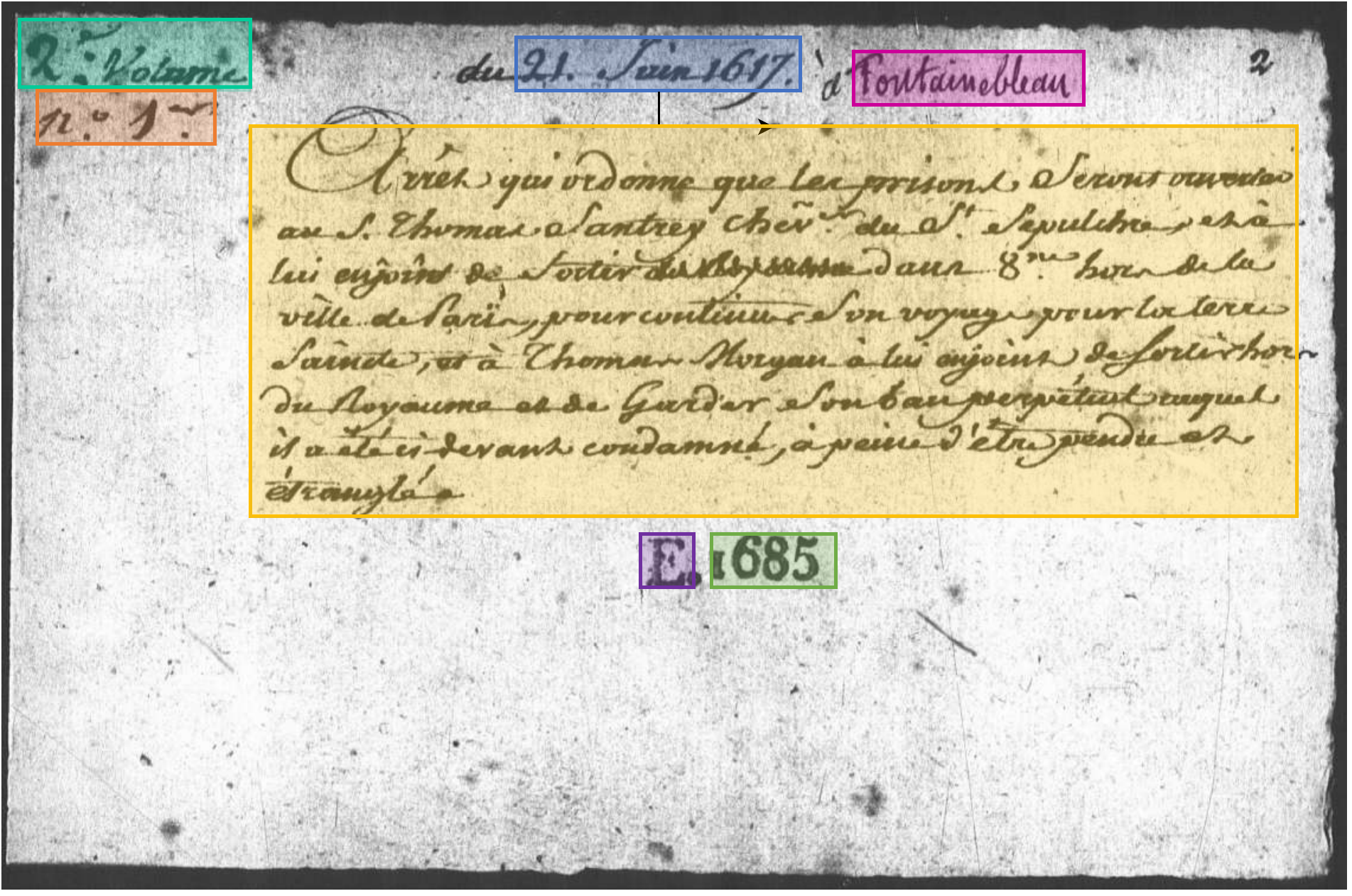}
         \caption{E series}
         \label{fig:serie_e}
     \end{subfigure}     
     \begin{subfigure}[b]{0.49\textwidth}
         \includegraphics[height=0.6\textwidth, center]{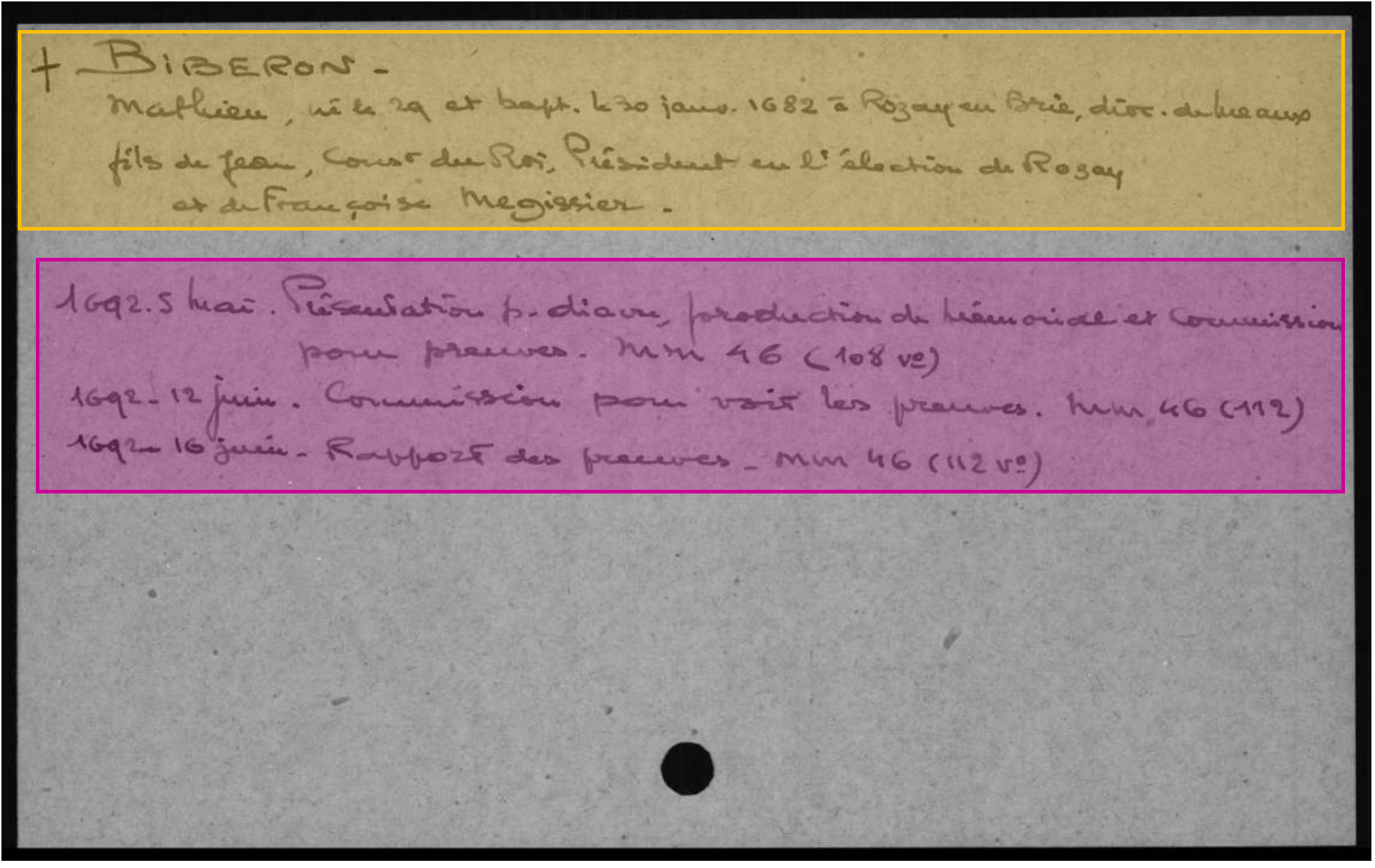}
         \caption{M series}
         \label{fig:serie_m}
     \end{subfigure}\vspace{1em}\\\vspace{1em}
     \hfill
     \begin{subfigure}[b]{0.49\textwidth}
         \includegraphics[height=1.3\textwidth, center]{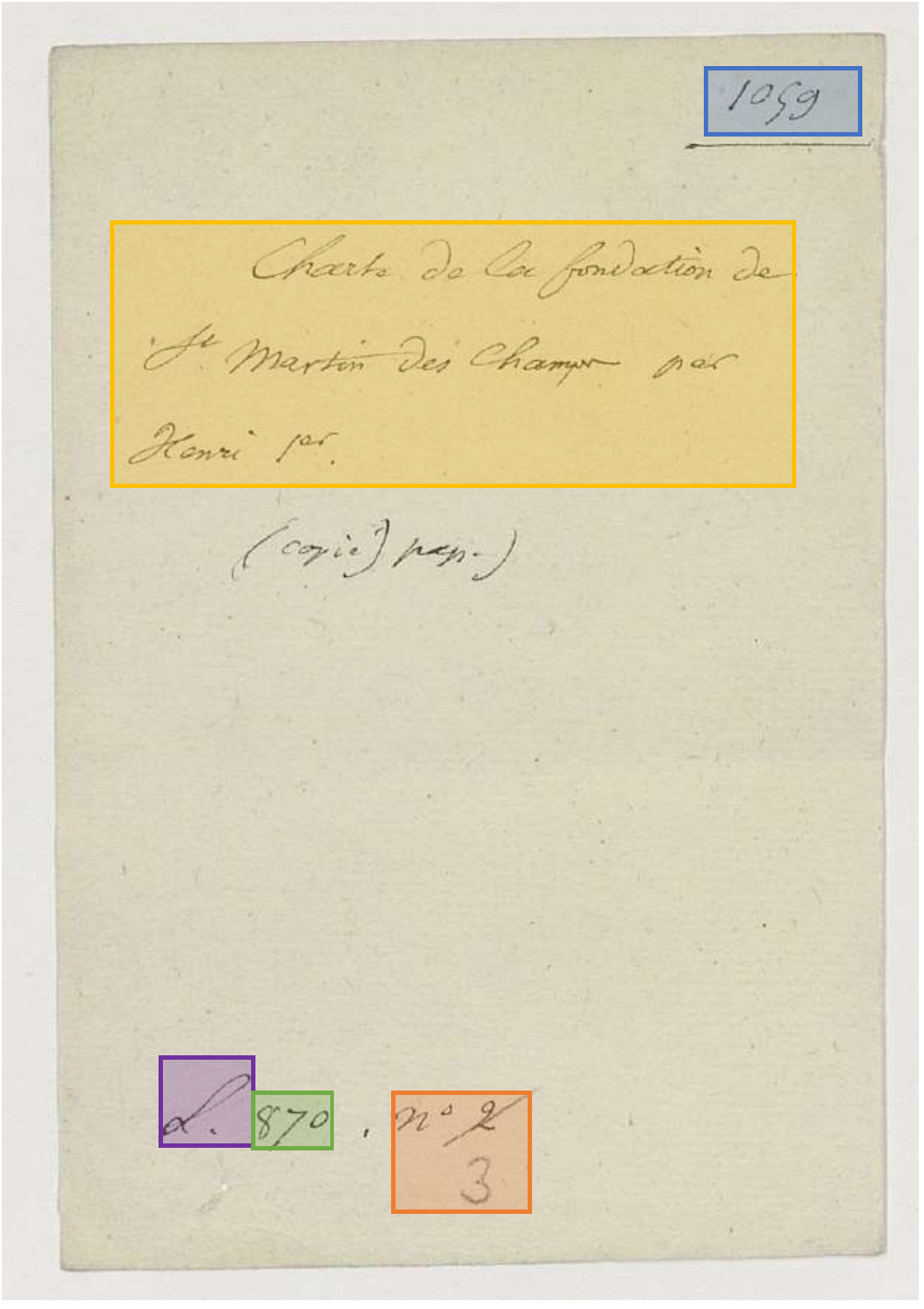}
         \caption{L series}
         \label{fig:serie_l}
     \end{subfigure}
     \begin{subfigure}[b]{0.49\textwidth}
         \includegraphics[height=1.3\textwidth, center]{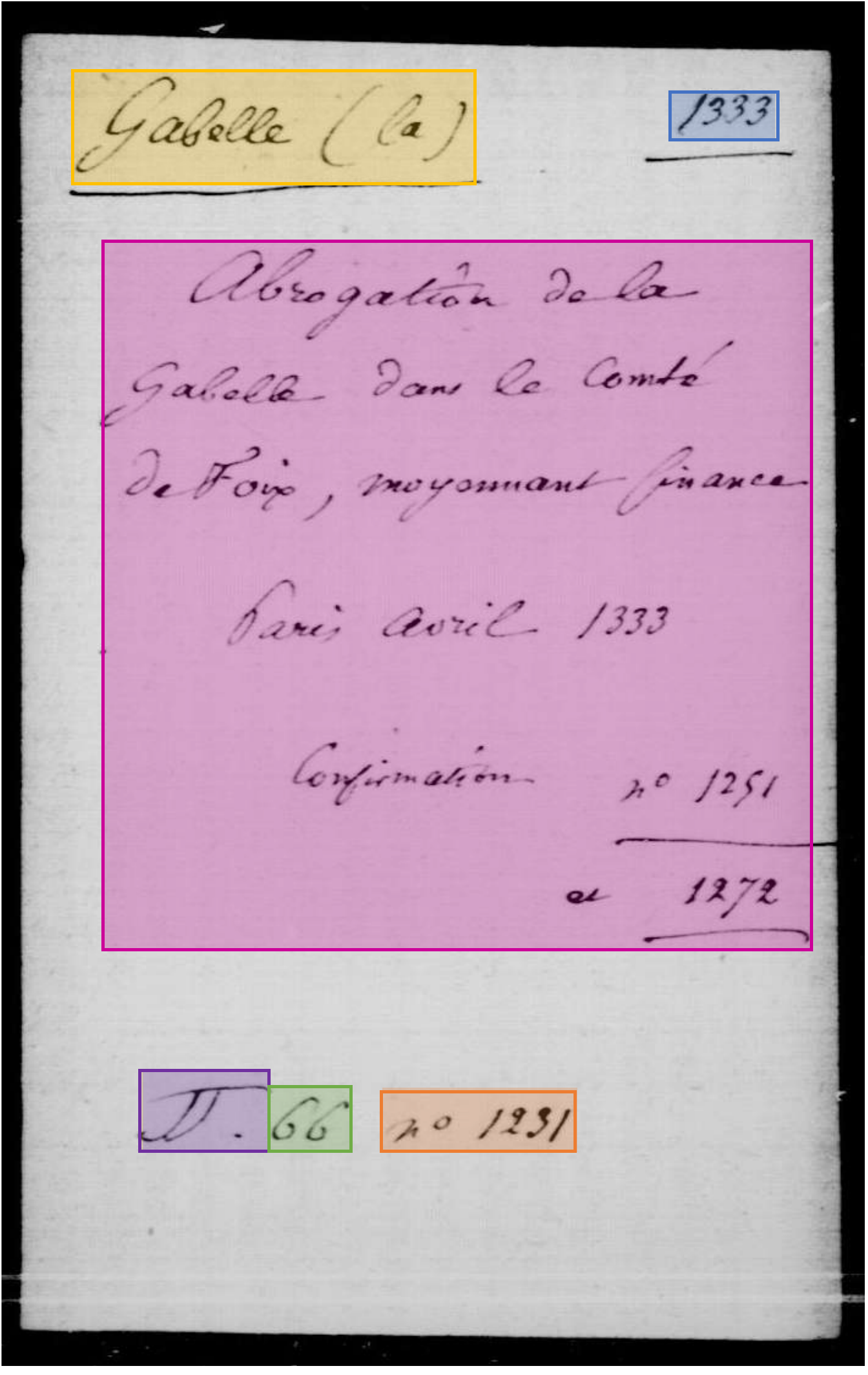}
         \caption{Douët d'Arcq (record of a JJ series document)}
         \label{fig:serie_m}
     \end{subfigure}\vspace{1em}
     \hfill
     \begin{subfigure}[b]{0.49\textwidth}
         \includegraphics[height=0.55\textwidth, center]{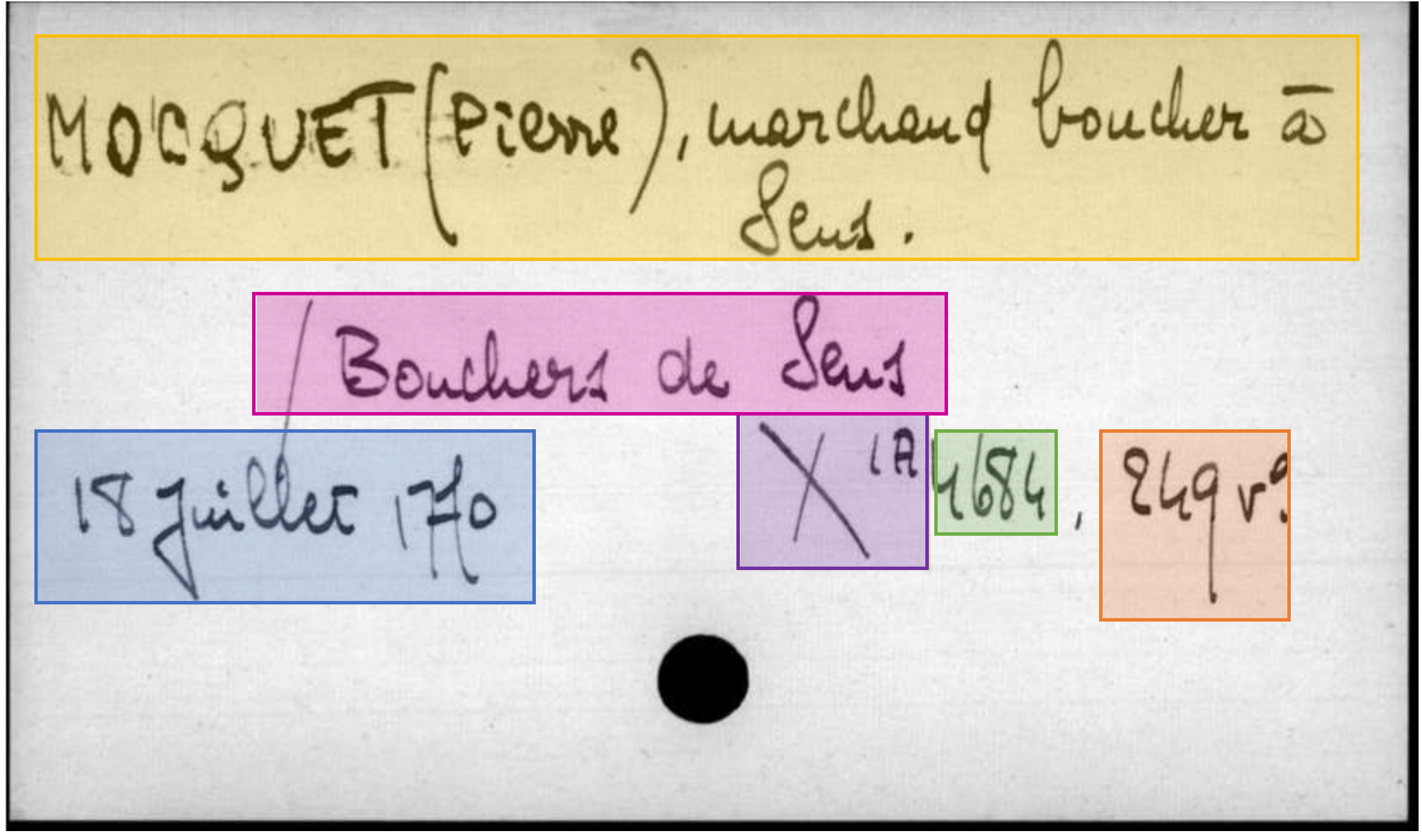}
         \caption{X1a series}
         \label{fig:serie_x}
     \end{subfigure}
     \begin{subfigure}[b]{0.49\textwidth}
         \centering
         \includegraphics[height=0.55\textwidth, center]{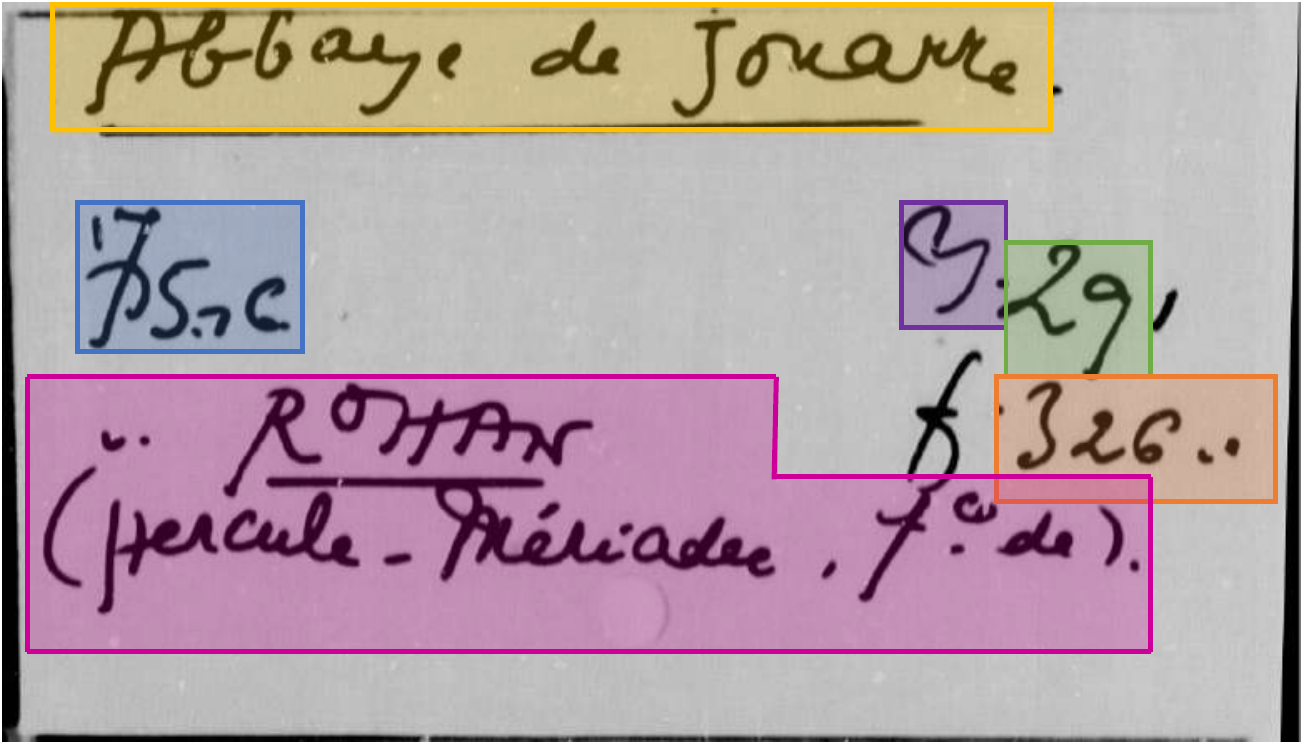}
         \caption{Y series}
         \label{fig:serie_y}
     \end{subfigure}
     \hfill
        \caption{Illustration of the different series of the SIMARA dataset and the information they contain. \textbf{Legend}: \cbox{blue} \texttt{date};  \cbox{yellow} \texttt{title};  \cbox{pink} \texttt{analysis}; \cbox{teal} \texttt{arrangement}; \cbox{purple} \texttt{serie}; \cbox{green} \texttt{article\_number}; \cbox{orange} \texttt{volume\_number}.}
        \label{fig:simara_datset}
\end{figure} 

In this section, we introduce the SIMARA dataset, illustrated in Fig. \ref{fig:simara_datset}. We describe the source of the documents, their usage, their structure and their content. 
The database containing both the images and the annotations is freely available at \url{https://zenodo.org/record/7868059}.

\subsection{Source of the documents} 
The documents of the SIMARA dataset were produced by the National Archives of France. The National Archives are a service 
in charge of preserving the archival heritage of the nation\footnote{\url{https://www.archives-nationales.culture.gouv.fr/en\_GB/web/guest/qui-sommes-nous}}. Their main missions are collecting the documents having both legal and historical interest, preserving them, whatever their medium is, and communicate them to the citizens, scholars and administrations.

The National Archives collect the documents from the current central administrations (presidency, ministries, public corporations) once their administrative usage has ended.
As a creation of the French Revolution, they also inherited the archives coming from the institutions that were abolished in 1789: mainly all the documents produced since the 6\textsuperscript{th} century by the royal institutions and the local Parisian administration, the charters of Parisian churches and abbeys and the Parisian notaries files.

Communicating all these documents (380 km of papers and 18 To of digital archives) would not be possible without sorting and describing them with metadata sets, so that the users can find the files that may interest them. 
These metadata sets are commonly named "finding aids" by archivists. The minimum metadata to identify archives are: a title, a date, an identifier (generally the unique reference of a box, a file, or even a document). They can be completed with other information such as dimensions, physical descriptions, and additional descriptive content. All this information is formalized in XML according to a specific DTD used worldwide: EAD\footnote{Encoded Archival Description. Official site: \url{https://www.loc.gov/ead/}}. They are published on websites to be requested through a search engine.

\subsection{Finding aids}  

The way in which the archives are described in finding aids can vary considerably. 
While the metadata of more recent documents are formalized directly in XML, the oldest series of the Middle and Modern Ages have handwritten finding aids, as they were first processed in the 19th century. The aim is to convert these into XML EAD files, but this poses several challenges:

\begin{itemize}
    \item \textit{Heterogeneity}: they were produced from the end of the 18\textsuperscript{th} century (sometimes even before the French Revolution) to the mid 20\textsuperscript{th} century. This involves a huge discrepancy in layout, content, language, and handwriting.
    \item \textit{Lack of standardization}: they were created before the digital age, in periods when archival description was not as codified as it is today, so the way in which content is presented varies greatly from one finding aid to another.
    \item \textit{Inaccessibility}: these finding aids were only digitized into image format (grayscale, 300 dpi, sometimes from microfilm or microfiche) in 2017-2019, with the aim to make their digital conversion easier, but they are not accessible to the public, neither in the reading room nor on the Internet.
    \item \textit{Quantity}: the index files are estimated to consist of around 800,000 paper cards. Using traditional methods, the digital conversion of these documents would take several years: manually typing the content and then manually encoding it into XML is a lengthy process that can be daunting.

\end{itemize}
The result of these considerations was the need for the use of automatic processes for the conversion of this material into digital data. 

\begin{table}[thb]
\centering
    \caption{Description of the information contained in the different series of the SIMARA dataset. }
    \label{tab:serie_description}
    \begin{tabular}{lp{0.82\textwidth}}
    \toprule
    \textbf{Serie} \phantom{--} & \textbf{Information to extract} \\
    \midrule
    E &  - Date (\texttt{date})\\ 
     & - Analysis of the decision (\texttt{title})\\ 
     &  - 1st part of the unique identifier (\texttt{serie}, \texttt{article\_number}) \\ 
     &  - 2nd part of the unique identifier (\texttt{volume\_number})\\ 
     &  - Original reference (\texttt{arrangement})\\ 
     &  - Number of the card (\texttt{analysis})\\
     \midrule
    L  & - Year of the charter (\texttt{date}) \\
     &  - Main analysis of the charter (\texttt{title}) \\
     & - Further details (\texttt{analysis}) \\
     & - Reference of the charter (\texttt{serie}, \texttt{article\_number}, \texttt{volume\_number})\\
     \midrule
     M & - Name and information on the person (\texttt{title}) \\
     &  - Description and proof of nobility (\texttt{analysis}) \\ 
     \midrule
    X1a & - Name of the first party (\texttt{title})\\
     &  - Name of the second party (\texttt{analysis})\\
     &  - Date of the trial (\texttt{date})\\
     &  - Reference of the judgement (\texttt{serie}, \texttt{article\_number}, \texttt{volume\_number})\\ 
    \midrule
    Y & - Name of the person  (\texttt{title})\\
     &   - Information about the person (\texttt{analysis})\\ 
     &  - Date (\texttt{date})\\
     &  - Reference (\texttt{serie}, \texttt{article\_number}, \texttt{volume\_number})\\ 
     \midrule
    Douet d'Arcq & - Heading of the record  (\texttt{title})\\
     &  - Complements and information (\texttt{analysis})\\ 
     & - Date (\texttt{date})\\
     & - Reference (\texttt{serie}, \texttt{article\_number}, \texttt{volume\_number})\\ 
    \bottomrule
\end{tabular}
\end{table}

\subsection{Description of the handwritten finding aids}
In this section, we describe the different series of finding aids included in the SIMARA datasetand the difficulties they present. Table \ref{tab:serie_description} summarizes the fields to be extracted in each series. We have also highlighted these fields for each series in Figure \ref{fig:simara_datset}.

\subsection*{E series ("King's councils")}

\textit{Type of archives:} registers in which the decisions of the King’s councils were recorded in the 17\textsuperscript{th} and 18\textsuperscript{th} centuries.\\
\textit{Finding aid:} 40~480 index cards from the 18\textsuperscript{th} century, containing an analysis of each decision of the King’s councils. This series is difficult to read due to the old French script and language.  


\subsection*{L series ("Spiritual life of churches and abbeys")}

\textit{Type of archives:} medieval charters of the Paris church of Saint-Martin-des-Champs.\\
\textit{Finding aid:} 623 index cards containing an analysis of each charter. They were written in the mid 19\textsuperscript{th} century. The main difficulties lie in the heterogeneous layout of the information and recent corrections of the identifiers (the old 19th identifier was crossed out with a pencil and corrected by a contemporary hand).

\subsection*{M series ("Proofs of nobility")}

\textit{Type of archives:} documents relating to the Knights of Malta and containing evidence of their nobility.\\\
\textit{Finding aid:} 4~847 cards arranged in alphabetical order. They were written around the 1950s. These documents do not present any particular difficulties.





\subsection*{X1a series ("Parliament")}

\textit{Type of archives:} registers in which were recorded the judgements from the Parliament, which was the highest court of justice.\\
\textit{Finding aid:} 101~036 cards containing an analysis of 18\textsuperscript{th} century trials involving two parties. These cards are sorted in alphabetic order by the name of the parties. They were written in the 20\textsuperscript{th} century, and do not present any particular difficulties.



\subsection*{Y series ("Châtelet")}

\textit{Type of archives:} notarial deeds of the 18\textsuperscript{th} century recorded in the Châtelet, which was a Parisian institution where the notaries had to report all the transactions they had recorded.\\
\textit{Finding aid:} 61~878 cards containing an analysis of the notarial deeds. The cards are sorted in alphabetical order by the name of the person who obtained the deed. They were written in the 1950's. The main difficulties are the poor digitization quality, the handwriting style, the heterogeneous layout as well as the density of the text.



\subsection*{Douët d'Arcq file}

\textit{Type of archives:} unlike the previous archives, several series are indexed by this file, mainly the "historical series" (J, K, L, M) which mainly correspond to the oldest and most prestigious documents (medieval royal archives, charters from Parisian abbeys and churches, University of Paris, Templars and Hospitallers…). Different kind of documents (charters, registers, accounts) are covered.\\
\textit{Finding aid:} 118~093 cards created in the mid 19\textsuperscript{th} century, and divided into three subsets: a file for personal names, another one for geographical names and a last one ordered by subjects. We only focus on the latter (6~284 cards). 



\subsection{Finding aids and original documents}
\begin{figure}[p]
    \begin{subfigure}[b]{0.45\textwidth}
    \centering
    \includegraphics[width=5cm]{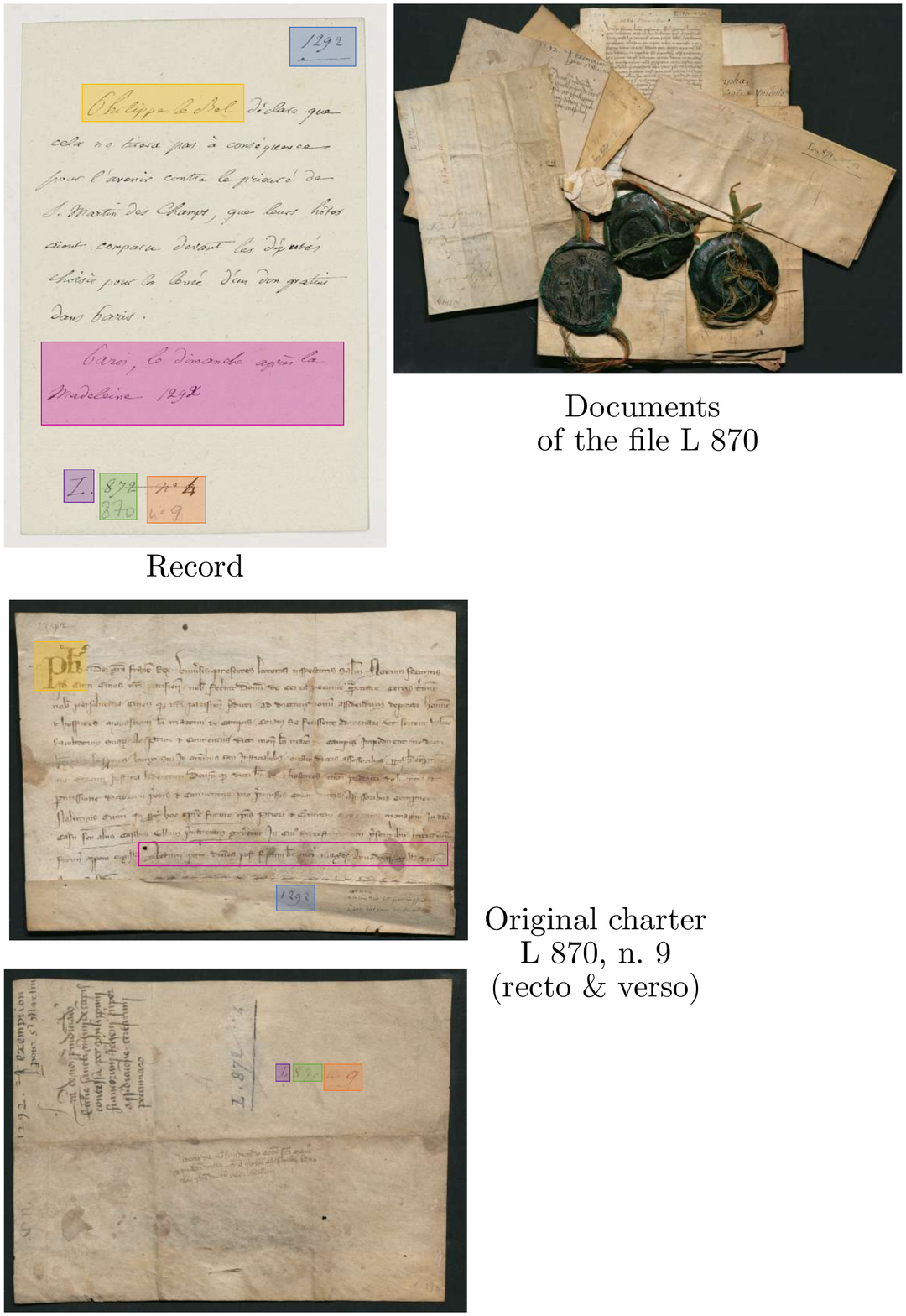}
    \caption{L series}
    \label{fig:l_series_document}
     \end{subfigure}
         \begin{subfigure}[b]{0.45\textwidth}
    \centering
    \includegraphics[width=5cm]{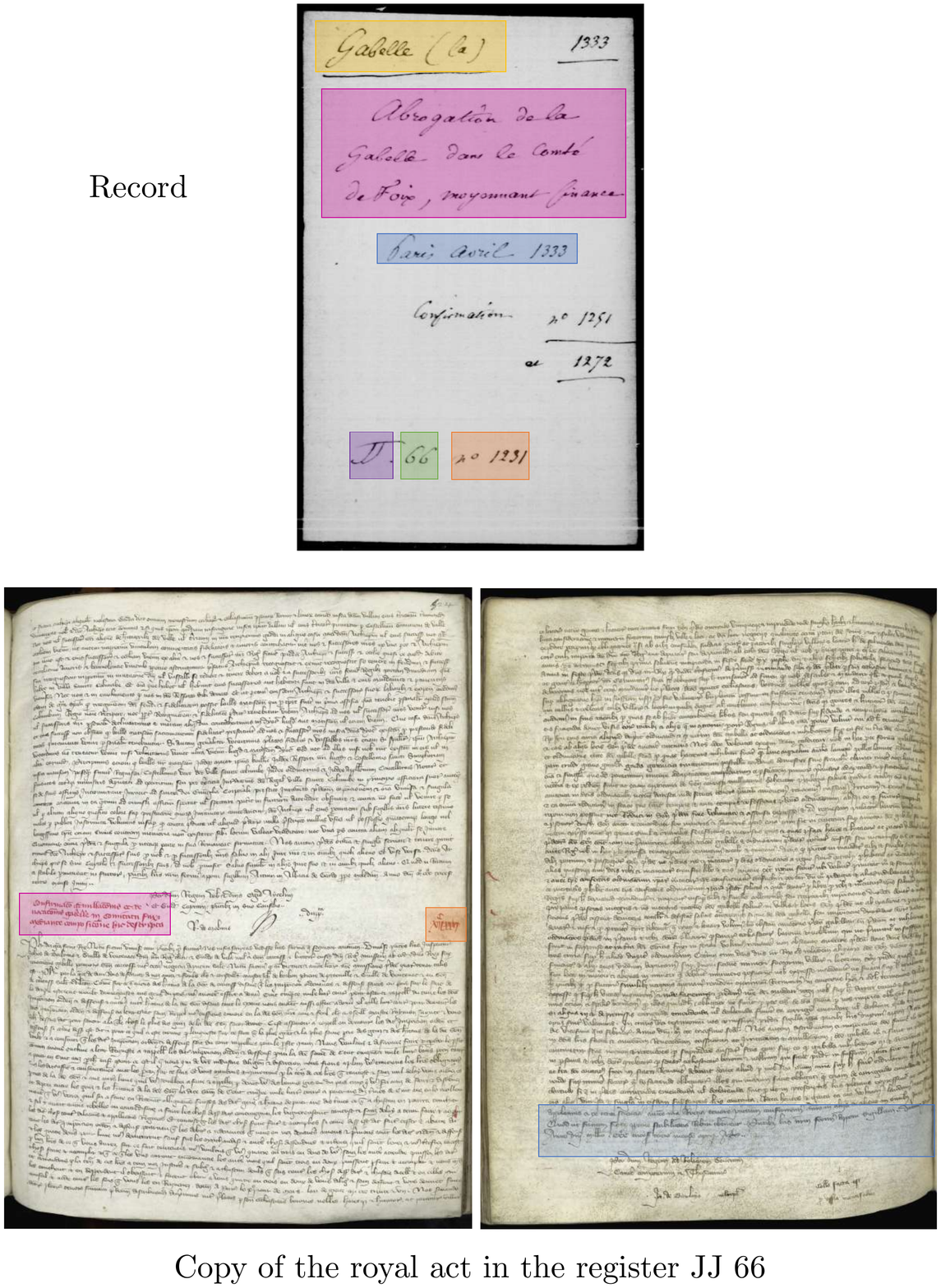}
    \caption{Douët d'Arcq file}
    \label{fig:dda_series_document}
     \end{subfigure}
    \begin{subfigure}[b]{0.45\textwidth}
    \centering
    \includegraphics[width=5cm]{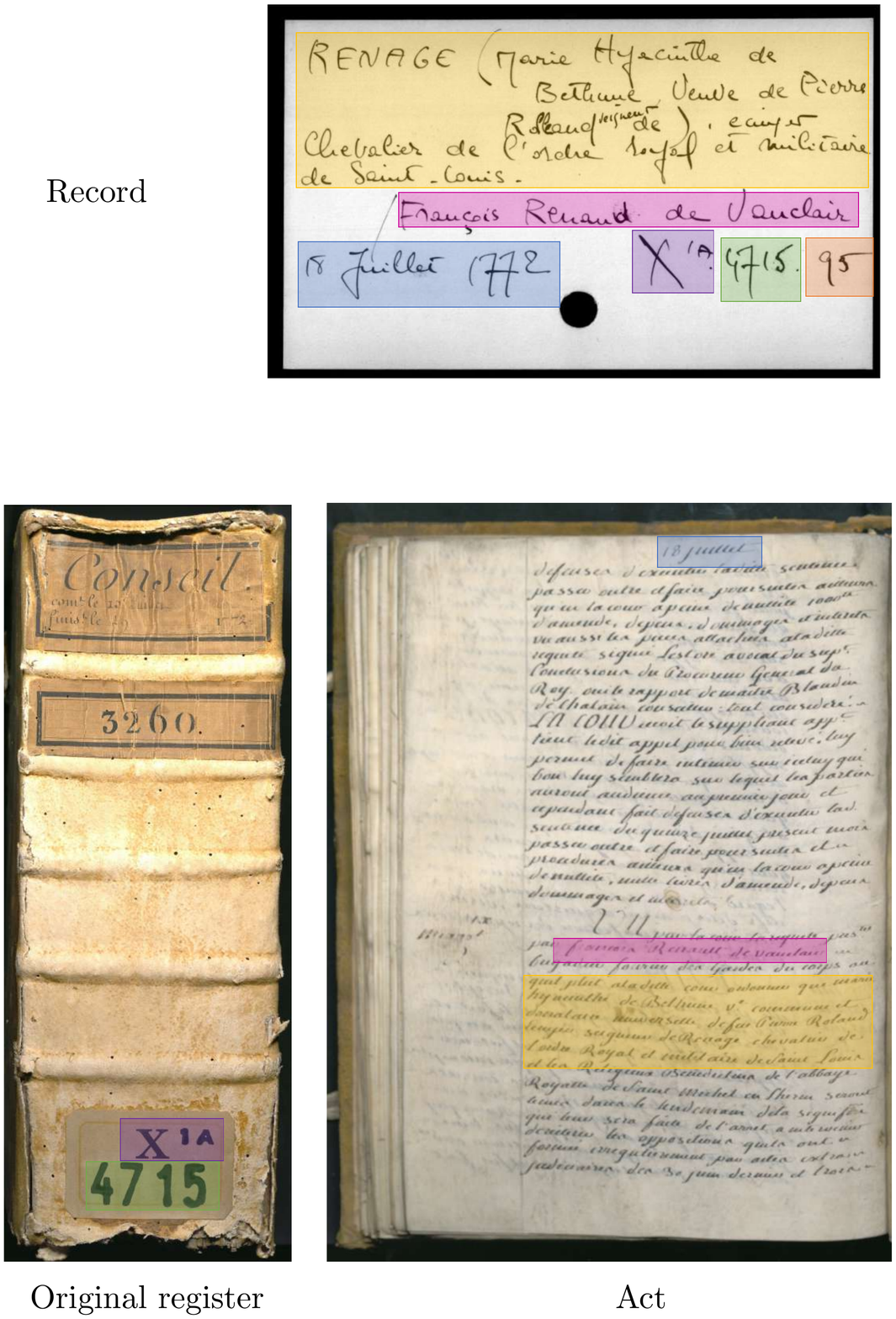}
    \caption{X series}
    \label{fig:x_series_document}
     \end{subfigure}
    \begin{subfigure}[b]{0.5\textwidth}
    \centering
    \includegraphics[width=5cm]{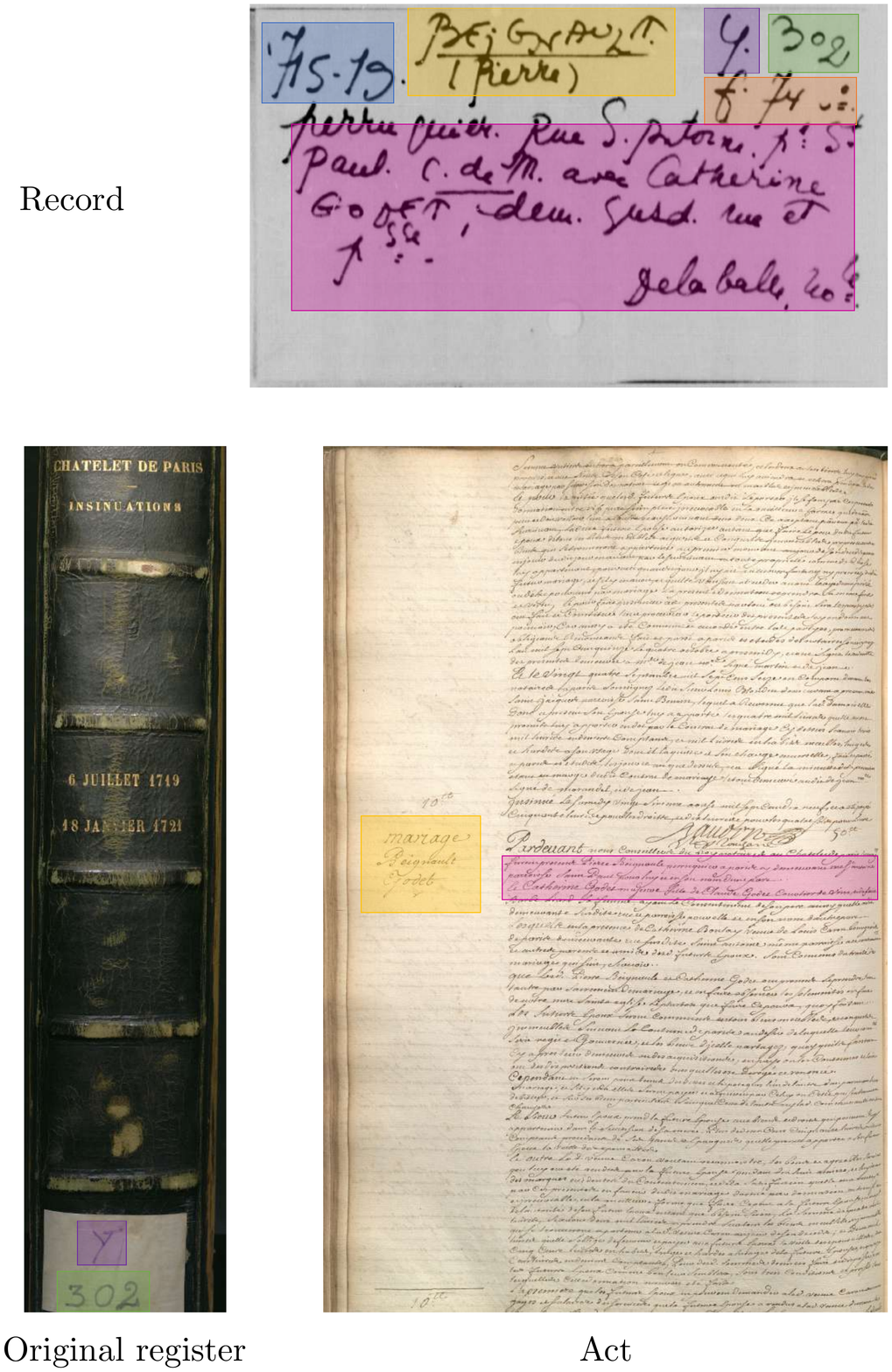}
    \caption{Y series.}
    \label{fig:y_series_document}
     \end{subfigure}
     \caption{Sample of records with the original document for four series.}
     \label{fig:record_and_docs}

\end{figure}

To give the reader a clearer idea of what finding aids and original records are, Figure \ref{fig:record_and_docs} shows examples of finding aids and reproductions of the original documents they describe. 

We have highlighted the main categories of information in the records, as shown in Figure \ref{fig:simara_datset}, and also where some of this information was found in the original documents. 

\subsection{Production of the annotated data}

The approach chosen for the generation of the annotated data of the SIMARA dataset was different from the traditionally used approaches. Usually, as processing chains are composed of different models for line detection, handwriting recognition and named entity extraction, it is necessary to generate specific annotations for each model: line positions on the image, transcribed text lines and entities positioned on the text. Each of these types of annotations requires a specific interface, and this interface is also different from the interface used to validate the model predictions in a production phase. The consequence of these multiple interfaces and multiple types of training data is that the data validated in production cannot be used to improve the models. To re-train the models, it is necessary to repeat annotation campaigns, in parallel with the validation tasks, which leads to a duplication of tasks. The approach chosen in this work is radically different: the annotation interface used to generate the first data needed to train the models is the same as the one used to validate the data in production. In this way, all production data can be used to iteratively improve the models.

\begin{figure}[ht]
    \centering
    \includegraphics[width=12cm]{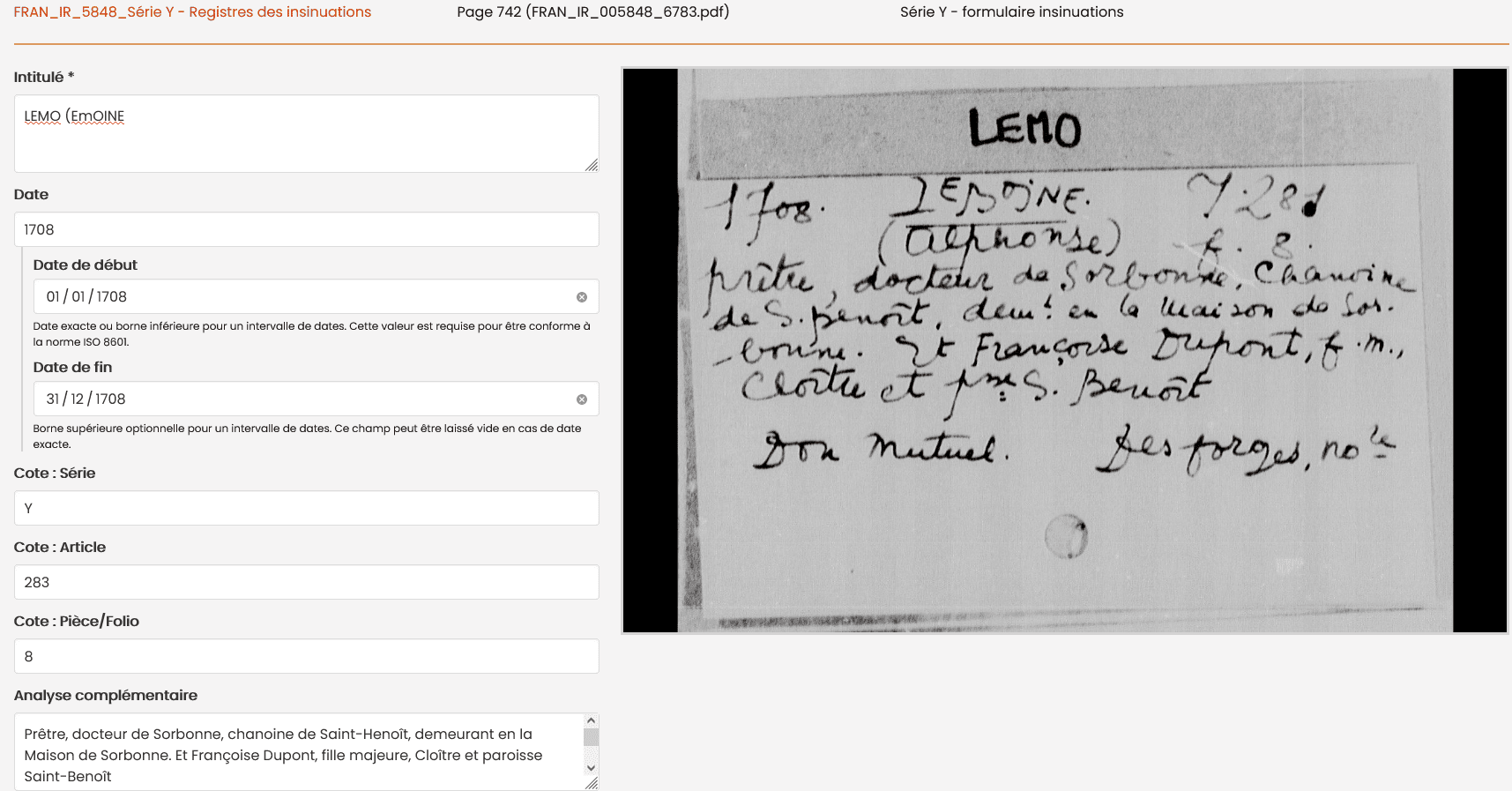}
    \caption{Annotator interface: the same interface is used to produce the ground-truth data and to validate the suggestions of the model in the production phase.}
    \label{fig:interf_simara_9}
\end{figure}

The originality of the SIMARA dataset is that the ground-truth was easily created by filling a form designed by the archivist. Each annotator was in charge of typing the ground-truth in the form by searching the information on the image of the finding aid. The annotation interface is shown on Figure \ref{fig:interf_simara_9}.

Once the models are trained, they are used in the production phase to pre-fill the form with suggestions. In this phase, the task of the annotators is different since they have to read, confirm or correct the information suggested in each field. This step is certainly the most time-consuming in the process, but it is crucial since the corrections made by the annotator are used to improve the information extraction model. The validation interface is the same as for the ground-truth generation.

The quality of the transcription was ensured by a process of double validation in case of doubt: when the annotators were uncertain about a piece of information, they could indicate it to an archivist for verification and validation. 

\section{Experimental results}
In this section, we report the results of the experiments conducted on the SIMARA dataset with our proposed system for full page key-value information extraction.

\subsection{Datasets}

Our training dataset evolved during the course of the project, as more and more documents were annotated and became available for training. 
As we trained models at different stages of the annotation procedure, we trained three models on different splits:
\begin{itemize}
    \item \textit{Split-v1}: 780 images for training, 50 for validation and  50 for testing; 
     \item \textit{Split-v2}: 3659 images for training, 783 for validation and 784 for testing; 
      \item \textit{Split-v3}: 3778 images for training, 811 for validation and 804 for testing. 
\end{itemize}

\begin{table}[th]
    \centering
    \caption{Statistics for the \textit{Split-v3} split}
    \label{tab:splitv3}
    \begin{subtable}{0.7\linewidth}
        \centering
        \caption{Number of images for each series}
        \begin{tabular}{lcccr}
            \toprule
            \textbf{Series} & \textbf{Train} & \textbf{Validation} & \textbf{Test} & \textbf{Total (\%)} \\
            \midrule
            E series & 322 & 64 & 79 & 8.6 \\
            L series & 38 & 8 & 4 & 0.9 \\
            M series & 128 & 21 & 27 & 3.3 \\
            X1a series & 2209 & 491 & 469 & 58.8 \\
            Y series & 940 & 205 & 196 & 24.9\\
            Douët s'Arcq series & 141 & 22 & 29 & 3.5 \\
            \midrule
            Total & 3778 & 811 & 804 & 100.0\\
        \bottomrule
        \end{tabular}
    \end{subtable}
    \vfill
    \begin{subtable}{0.7\linewidth}
        \centering
        \caption{Number of entities (word count)}
        \begin{tabular}{lcccr}
            \toprule
            \textbf{Entities} & \textbf{Train} & \textbf{Validation} & \textbf{Test} & \textbf{Total (\%) }\\
            \midrule
            \texttt{date} & 8406 & 1814 & 1799 & 10.4 \\
            \texttt{title} & 35531 & 7495 & 8173 & 44.4 \\
            \texttt{serie} & 3168 & 664 & 676 & 3.9 \\
            \texttt{analysis} & 25988 & 5130 & 5602 & 31.8 \\
            \texttt{volume\_number} & 3913 & 808 & 813 & 4.8  \\
            \texttt{article\_number} & 3181 & 665 & 678 & 3.9 \\
            \texttt{arrangement} & 644 & 122 & 153 & 0.8\\
            \midrule
            Total & 80831 & 16698 & 17894 & 100.0\\
        \bottomrule
        \end{tabular}
    \end{subtable}
\end{table}

\subsection{Key-value information extraction model}

We train a model based on the DAN architecture \cite{DAN} for key-value information extraction from full pages on SIMARA. DAN\footnote{\url{https://github.com/FactoDeepLearning/DAN}} is an open source attention-based Transformer model for handwritten text recognition that can work directly on pages. It is trained with the cross-entropy loss function, and its last layer is a linear layer with a softmax activation to compute probabilities for each character of the vocabulary. 
We address the task of key-value information extraction by encoding each field with a special token located at the beginning of each section, as illustrated in Table \ref{tab:simara_dataset}. A complete description and evaluation of this model is presented in \cite{tarride2023b}.

\begin{table}[th]
    \centering
    \caption{Example of transcription used to train our baseline model. Special tokens are localized at the beginning of each text section to characterize the corresponding field. Note that the fields always appear in the same order in transcriptions, even if it does not correspond to the reading order on the images.}
    \label{tab:simara_dataset}
    \begin{tabular}{p{0.8\textwidth}}   
    \begin{center}
    \raisebox{-\totalheight}{\includegraphics[width=0.7\textwidth]{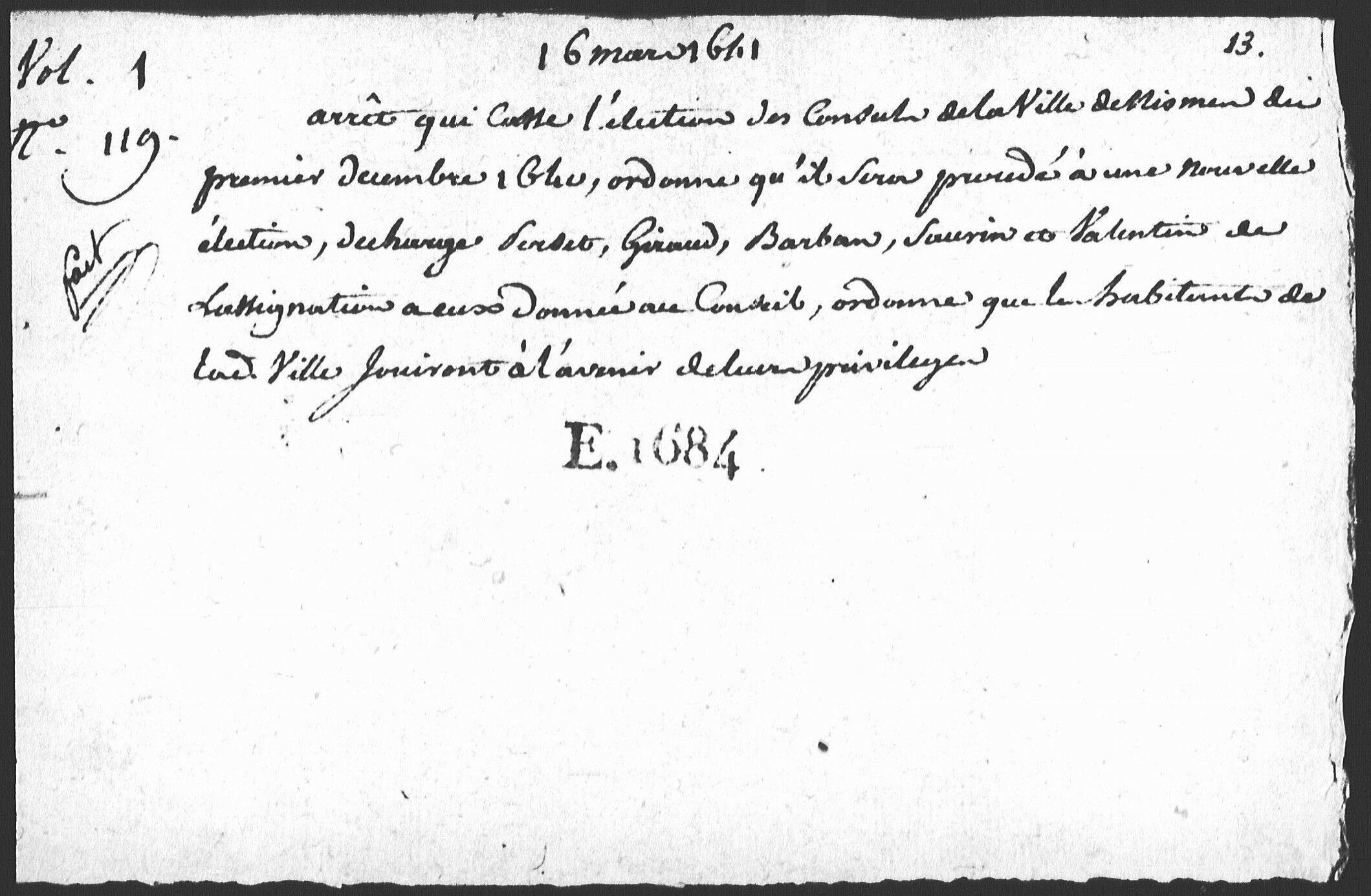}} 
    \end{center}
    \\
    \toprule
        Transcription \\
        \midrule
        \texttt{\color{orange}<reference\_number>\color{black}\orangeuline{n° 119}\color{black} \color{blue}<date>\color{black}\blueuline{16 mars 1641} \color{pink}<analysis>\color{black}\pinkuline{13} \color{yellow}<title>\color{black}\yellowuline{Arrêt qui casse l'élection des consuls de la ville de Nîmes du premier décembre 1640, ordonne qu'il sera procédé à une nouvelle élection, décharge Perset, Giraud, Barban, Saurin et Valentin de l'assignation à eux donnée au Conseil, ordonne que les habitants de ladite ville jouiront à l'avenir de leurs privilèges.} \color{purple}<serie>\color{black}\purpleuline{E} \color{green}<article\_number>\color{black}\greenuline{1684} \color{teal}<arrangement>\color{black}\tealuline{Volume 1}} \\
        \bottomrule

    \end{tabular}

\end{table}

\subsection{Metrics}

\paragraph{HTR metrics:}
the quality of handwriting recognition is evaluated using the standard Character Error Rate (CER) and Word Error Rate (WER). The full text is evaluated, but named entities are ignored  for this kind of the evaluation. 

\paragraph{NER metrics:}
we use the Nerval\footnote{\url{https://gitlab.com/teklia/ner/nerval}} evaluation toolkit to evaluate Named Entity Recognition results. Nerval is able to deal with noisy text by aligning the automatic transcription with the ground truth at character level. Predicted and ground-truth entities are considered a match if their label is similar and their edit distance below a threshold, set to 30\% in our experiments. 
From this alignment, precision, recall and F1-score are computed to evaluate Named Entity Recognition. 

\subsection{Results}

In this section, we present the performance reach by our baseline model for key-value information extraction on SIMARA. First, we study the impact of the training dataset size, then we detail the results obtained with the best model.

\subsubsection{Impact of training dataset size}

We trained three models on three different data splits, using an increasingly bigger training set. Tab. \ref{tab:models_by_split} compares the results obtained with different versions of the training set. As expected, the results improve as the training set gets larger. It is interesting to note that even a small increase in the amount of training data, e.g. a hundred images between \textit{Split-v2} and \textit{Split-v3}, improves the recognition rate in our case by one CER point.

\begin{table}[t]
    \centering
    \caption{Results of the three models on the test set.}
    \label{tab:models_by_split}
    \begin{tabular}{lcccr}
        \toprule
        \textbf{Model name} \phantom{-}& \textbf{Split} & \textbf{CER (\%)} & \textbf{WER (\%)} & \textbf{N images} \\
        \midrule
        \textit{Model-v1} & \textit{Split-v1} & 12.08 & 22.41 & 50 \\
        \textit{Model-v2} & \textit{Split-v2} & 7.39 & 15.07 & 784 \\
        \textit{Model-v3} & \textit{Split-v3} & \textbf{6.46} & \textbf{14.79} & 804 \\
        \bottomrule
    \end{tabular}
\end{table}

\subsubsection{Detailed HTR results for the best model}

We provide detailed results for the best model (\textit{Model-v3}) in Table \ref{tab:htr_results_modelv3}. Character and Word Error Rates are computed on the test set for each series of documents. 
The \textit{L series} is the best recognized, although it is not well represented in the training set. 
It should be noted that this series only has 4 documents in the test set, which limits the statistical interpretation of this observation.
The most difficult series is Douët d’Arcq, with a WER close to 25\%. This can be explained by the fact that this series has a unique layout and few examples in the training set. 
The other most difficult series are the Y series (high CER) and the E series (high WER) which, although highly represented in the training set, present some difficulties in terms of layout and language respectively.
Overall, text recognition results are very acceptable, with a WER below 15\%. 

\begin{table}[th]
    \centering
    \caption{Detailed evaluation results for \textit{Model-v3}.}
    \label{tab:splitv3}
    \begin{subtable}{0.95\linewidth}
        \centering
    \caption{Character and Word Error Rates are computed for each serie on the test set. Special tokens encoding each field are removed at this stage of the evaluation. }
    \label{tab:htr_results_modelv3}
    \begin{tabular}{lccrrr}
        \toprule
        \textbf{Series} & \textbf{CER (\%) }&\textbf{ WER (\%)} & \textbf{N images} \\
        \midrule
        E series & 6.89 & 18.73 & 79 \\ 
        L series & \textbf{3.33} & \textbf{9.35} & 4 \\
        M series & 7.70 & 15.87 & 27 \\
        X1A series & 4.97 & 11.16 & 469 \\
        Y series & 8.23 & 16.19 & 196 \\
        Douët d’Arcq & 10.24 & 24.97 & 29 \\
        \midrule
        Total & 6.46 & 14.79 & 804 \\
    \bottomrule
    \end{tabular}
    \end{subtable}
    \vfill
    \begin{subtable}{0.99\linewidth}
        \centering
        \caption{NER precision, recall and F1-score are computed for each field on the test set. }
        \label{tab:ner_results_modelv3}
        \begin{tabular}{lcccccr}
            \toprule
            \textbf{Tag}          & \textbf{Predicted} & \textbf{Matched} & \textbf{Precision (\%)} & \textbf{Recall (\%)} & \textbf{F1 (\%)} & \textbf{N entities} \\
            \midrule
            \texttt{date}         & 1808      & 1784    & 98.67         & \textbf{99.17}      & \textbf{98.92}  & 1799    \\
            \texttt{title}        & 8059      & 7650    & 94.92         & 93.60      & 94.26  & 8173    \\
            \texttt{serie}        & 677       & 664     & 98.08         & 98.22      & 98.15  & 676     \\
            \texttt{analysis}     & 5449      & 5248    & 96.31         & 93.68      & 94.98  & 5602    \\
            \texttt{article\_number}   & 672       & 664     & \textbf{98.81}         & 97.94      & 98.37  & 678     \\
            \texttt{volume\_number}  & 805       & 776     & 96.40         & 95.45      & 95.92  & 813     \\
            \texttt{arrangement}   & 206       & 119     & 57.77         & 77.78      & 66.30   & 153     \\
            \midrule
            Total          & 17676     & 16905   & 95.64         & 94.47      & 95.05  & 17894   \\
    \bottomrule
    \end{tabular}
    \end{subtable}
\end{table}

\subsubsection{Detailed NER results for the best model}

These documents are difficult to evaluate from a transcription point of view because the reading order of the machine does not necessarily coincide with that of the annotator. Therefore, a high error rate at character or word level does not necessarily indicate poor transcription quality, but rather differences in the reading order.

An evaluation of the detection of key-value information (named entity), allows to overcome the problems of reading order. The table \ref{tab:ner_results_modelv3} shows the results of automatic extraction with the \textit{Model-v3}, detailed by type of information. We note very good performances on numerical information such as date, serial number and item number. Textual information such as the title and the analysis are also extracted with very good rates. Only information that is not very present in the training set, such as the arrangement field, performs less well. In general, more than 95\% of the information is extracted correctly.

\section{Conclusion}

In this article, we have described SIMARA, a new public database consisting of images of digitised finding aids with the transcription of the information they contain. This database has allowed us to define a new type of task for automatic historical document processing systems. This task consists in the extraction of key-value information from document pages. 
We proposed a baseline system based on a Transformers neural network, allowing the extraction of this information in a single pass. The performances of this model constitute reference results, allowing to evaluate the progress of models which will be proposed to solve this task on the SIMARA database.

\bibliographystyle{splncs04}
\bibliography{mybibliography}
\end{document}